\documentclass[sigconf]{acmart}
\AtBeginDocument{%
  \providecommand\BibTeX{{%
    \normalfont B\kern-0.5em{\scshape i\kern-0.25em b}\kern-0.8em\TeX}}}

\setcopyright{acmcopyright}
\copyrightyear{2018}
\acmYear{2018}
\acmDOI{XXXXXXX.XXXXXXX}

\acmConference[Conference acronym 'XX]{Make sure to enter the correct
  conference title from your rights confirmation emai}{June 03--05,
  2018}{Woodstock, NY}
\acmPrice{15.00}
\acmISBN{978-1-4503-XXXX-X/18/06}

\usepackage{cleveref}

\begin{document}

\title{Sarcasm Detection using Hybrid Neural Network}

\author{Rishabh Misra}
\affiliation{%
  \institution{Twitter, Inc}
  \city{San Francisco}
  \country{USA}}
\email{r1misra@eng.ucsd.edu}

\author{Prahal Arora}
\affiliation{%
  \institution{Facebook AI}
  \city{New York}
  \country{USA}}
\email{prarora@eng.ucsd.edu}

\renewcommand{\shortauthors}{Misra and Arora}

\begin{abstract}
    Sarcasm has been an elusive concept for humans. Due to interesting linguistic properties, Sarcasm Detection has recently gained some attraction of the Natural Language Processing research community. However, the task of predicting sarcasm in a text remains a difficult one for machines as well, and there are limited insights into what makes a sentence sarcastic. Past studies mostly make use of Twitter-based datasets collected using hashtag-based supervision but such datasets are noisy in terms of labels and language - thus limiting the interpretability. To overcome these shortcomings, we introduce a new dataset which is a collection of news headlines from a sarcastic news website and a real news website. Utilizing this high-quality dataset, we further propose an interpretable Hybrid Neural Network architecture which provides insights into what actually makes sentences sarcastic. Through quantitative experiments, we show that the proposed model improves upon a strong baseline by $\sim$ 5\% in terms of classification accuracy. Lastly, we make the dataset as well as framework implementation publicly available to facilitate future research in this domain.
\end{abstract}





\maketitle

\section{Introduction}
There have been many studies on Sarcasm Detection in the past that have either used a small high-quality labeled dataset or a large noisy labeled dataset. In each type of scenario, the interpretability of sarcasm is limited by the access to large and high-quality dataset. One of the prominent works in this domain by Amir et. al. \cite{amir2016modelling} use a large-scale Twitter-based dataset collected using hashtag-based supervision. They propose to use a CNN to automatically extract relevant features from tweets and augment them with user embeddings to provide more contextual features during Sarcasm Detection. However, this work is limited in following aspects:
\begin{itemize}
    \item Twitter-based dataset used in the study was collected using hashtag-based supervision. As per various studies \cite{Liebrecht2013ThePS, joshi2017automatic}, such datasets have noisy labels. Furthermore, people use very informal language on Twitter which introduces sparsity in vocabulary and for many words, pre-trained embeddings are not available. Lastly, many tweets are replies to other tweets and detecting sarcasm in such cases require the availability of contextual tweets.
    \item The proposed framework is quite simplistic. Authors use CNN with one convolutional layer to extract relevant features from text which are then concatenated with (pre-trained) user embeddings to produce the final classification score. However, other studies like \cite{yin2017comparative} show that RNNs are more suitable for sequential data. Furthermore, authors propose a separate method to learn the user embeddings which means the model is not trainable end to end.
    \item There is no available qualitative analysis from the proposed framework to showcase what the model is learning and in which cases it is performing well.
\end{itemize}

We understand that detecting sarcasm requires understanding of common sense knowledge, without which the model might not actually understand what sarcasm is and may just pick up some discriminative lexical cues. This direction has not been addressed in previous studies to the best of our knowledge. Due to these limitations, it has been difficult to understand and interpret the elusive concept of Sarcasm. To tackle these challenges, we summarize our contributions in this work as follows:
\begin{itemize}
    \item We first describe a newly collected large-scale dataset for sarcasm detection that is superior in terms of labels and language as compared to previously available high-quality datasets in this domain.
    \item We propose an interpretable Hybrid Neural Network that outperforms a strong baseline by $\sim$ 5\% in terms of classification accuracy on the newly collected dataset.
    \item Lastly, we try to interpret the concept of sarcasm through the proposed model's attention module.
\end{itemize}

The rest of the paper is organized in following manner: in \cref{sec:sec2}, we describe the dataset collected by us to overcome the limitations of Twitter-based high-quality dataset. In \cref{sec:sec3}, we describe the network architecture of the proposed model. In \cref{sec:sec4} and \cref{sec:sec5}, we provide experiment details, results and analysis. To conclude, we provide few future directions in \cref{sec:sec6}.

\section{Dataset Collection} \label{sec:sec2}
\begin{figure}[t!]
    % \begin{subfigure} %[h]{0.49*\linewidth}
    \includegraphics[width=\linewidth]{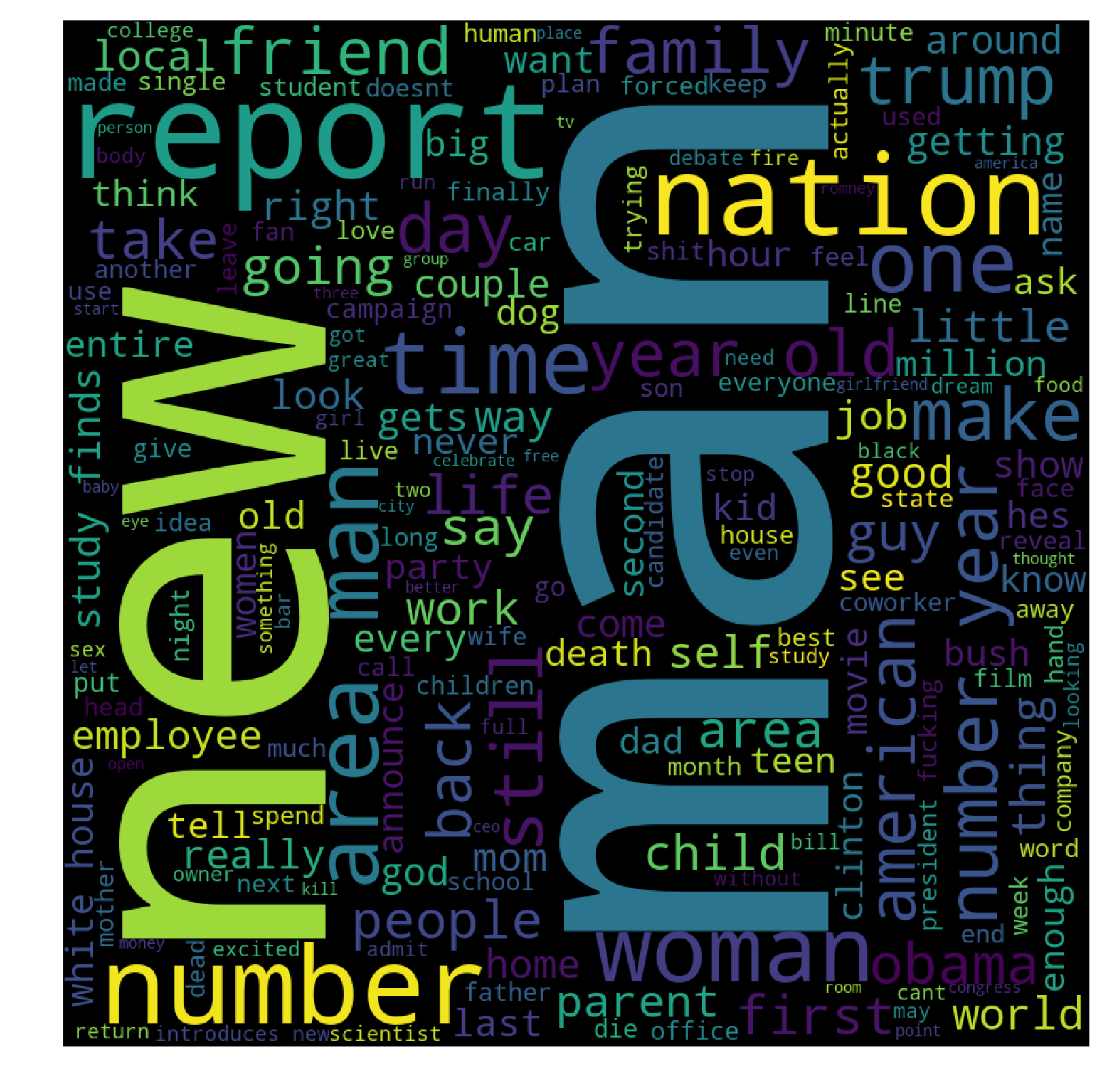}
    \caption{Wordcloud of sarcastic headlines.}
    \label{fig:fig1a}
    \vspace*{-5mm}
    %\caption{Effectiveness of metric learning approach in classifying infrequent labels.}
    %\label{fig:fig1a}
    % \end{subfigure}
% \hfill
%     \begin{subfigure} %[h]{0.49*\linewidth}
\end{figure}

\begin{figure}[h]
    \includegraphics[width=\linewidth]{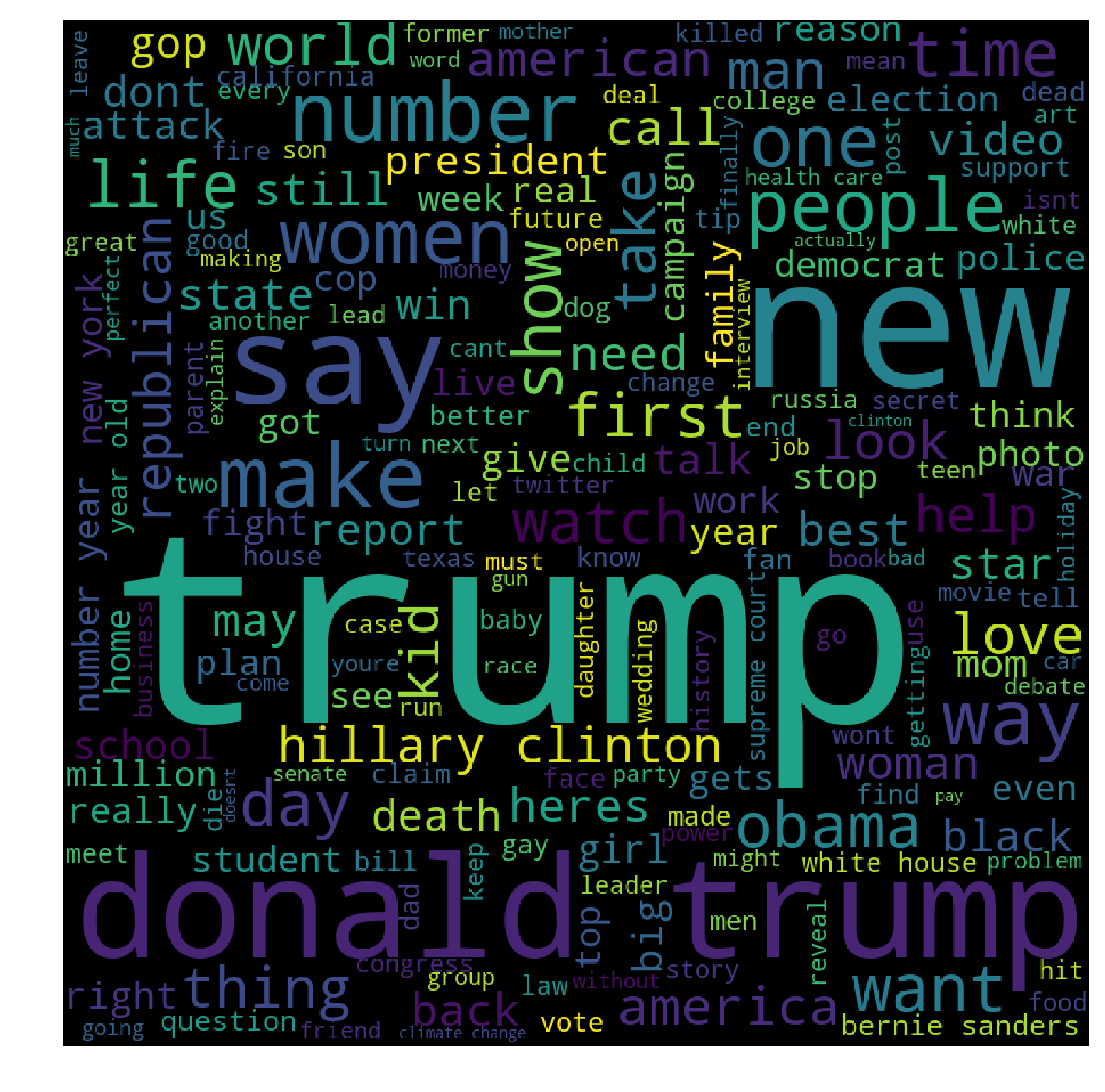}
    \vspace*{-5mm}
    %\caption{Average test AUC for products with given number of transactions in training data.}
    %\label{fig:fig1b}
    % \end{subfigure}
    \caption{Wordcloud of non-sarcastic headlines.}
    \label{fig:fig1b}
\end{figure}

To overcome the limitations related to label and language noise in Twitter-based datasets, we collected a \emph{News Headlines Dataset}\footnote{\url{https://rishabhmisra.github.io/publications}} from two news websites. \emph{TheOnion}\footnote{\url{https://www.theonion.com/}} aims at producing sarcastic versions of current events and we collect all the headlines from News in Brief and News in Photos categories (which are sarcastic). We collect real (and non-sarcastic) news headlines from \emph{HuffPost}\footnote{\url{https://www.huffingtonpost.com/}}. For exploring the text, we visualize the word clouds in \Cref{fig:fig1a} and \Cref{fig:fig1b} through which we can see the types of words that occur frequently in each category. The general statistics of this dataset along with Twitter-based dataset provided by Semeval challenge\footnote{\url{https://competitions.codalab.org/competitions/17468}} are given in \Cref{tab:table1}. We can notice that for News Headlines Dataset, the percentage of words not available in word2vec vocabulary is significantly less than Semeval dataset.

\begin{table}[h!]
  \centering
  \resizebox{\columnwidth}{!}{%
  \begin{tabular}{lrr}
  \toprule
  \textbf{Statistic/Dataset} & \textbf{Headlines} & \textbf{Semeval} \\
  \midrule
  \# Records & 26,709 & 3,000 \\
  \# Sarcastic records & 11,725 & 2,396 \\
  \# Non-sarcastic records & 14,984 & 604 \\
  \% word embeddings not available & 23.35 & 35.53 \\
  \bottomrule
\end{tabular}}
  \caption{General statistics of datasets.}
  \label{tab:table1}
\end{table}

To summarize, the \emph{News Headlines Dataset} has following advantages over the existing Twitter datasets:
\begin{itemize}
\item Since news headlines are written by professionals in a formal manner, there are no spelling mistakes or informal usage. This reduces the sparsity and also increases the chance of finding pre-trained embeddings.
\item Since the sole purpose of \emph{TheOnion} is to publish sarcastic news, we get high quality labels with much less noise as compared to Twitter-based datasets.
\item Unlike tweets which are replies to other tweets, the news headlines we obtained are self-contained. This would help us in teasing apart the real sarcastic elements. 
\end{itemize}

\section{Interpretable Neural Network}  \label{sec:sec3}
The original architecture of Amir et. al. \cite{amir2016modelling} takes pre-trained user embeddings (context) and tweets (content) as input and outputs a binary value for sarcasm detection. We tweaked this architecture to remove the user-context modeling path since the mention of sarcasm in this dataset is not dependent on authors but rather on current events and common knowledge. %it is difficult to obtain pre-trained user embeddings (or user context) for free text dataset scraped from the Internet. Hence, the dependency on user-context input is a major drawback for any such model. 
In addition to that, a new LSTM module is added to encode the left (and right) context of the words in a sentence at every time step. This LSTM module is supplemented with an Attention module to reweigh the encoded context at every time step. 

\begin{figure*}[!ht]
	\centering
    \includegraphics[width=0.8\linewidth]{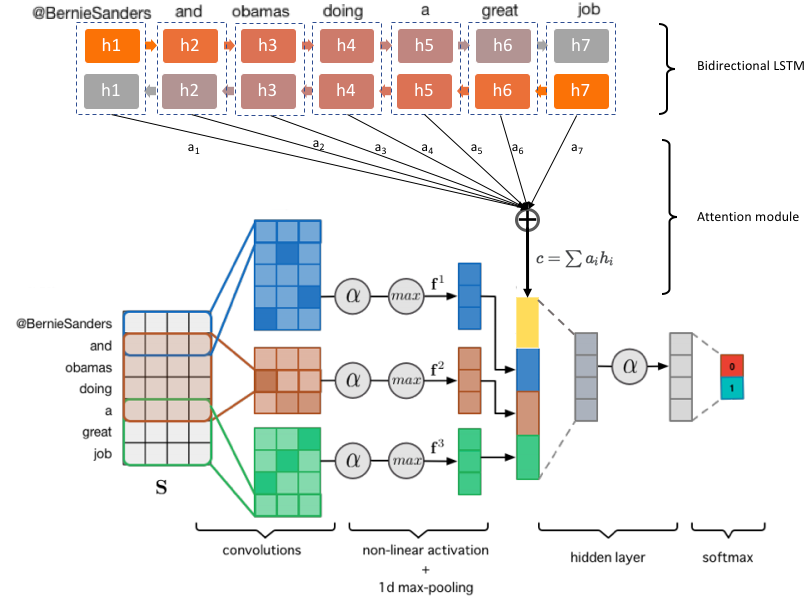}
    \caption{Interpretable Hybrid Neural Network Architecture}
    \label{fig:fig3}
\end{figure*}

We hypothesize that the sequential information encoded in LSTM module would complement the existing CNN module in the original architecture of \cite{amir2016modelling} which captures regular n-gram word patterns throughout the entire length of the sentence. We also hypothesize that attention module can really benefit the task at hand. It can selectively emphasize on incongruent co-occurring word phrases (words with contrasting implied sentiment). For example, in the sentence ``majority of nations civic engagement centered around oppressing other people", our attentive model can emphasize on occurrence of `civic engagement' and `oppressing other people' to classify this sentence as sarcastic. The detailed architecture of our model is illustrated in figure \ref{fig:fig3}. 

The LSTM module with attention is similar to the one used to jointly align and translate in a Neural Machine Translation task \citep{bahdanau2014neural}. A BiLSTM consists of forward and backward LSTMs. The forward LSTM calculates a sequence of forward hidden states and the backward LSTM reads the sequence in the reverse order to calculate backward hidden states. We obtain an annotation for each word in the input sentence by concatenating the forward hidden state and the
backward one. In this way, the annotation $h_j$ contains the summaries of both the preceding words and the following words. Due to the tendency of LSTMs to better represent recent inputs, the annotation at any time step will be focused on the words around that time step in the input sentence.  Each hidden state contains information about the whole input sequence with a strong focus on the parts surrounding the corresponding input word of the input sequence. The context vector $\mathbf{c}$ is, then, computed as a weighted sum of these annotations. 
\vspace{-3mm}
$$ \mathbf{c} = \sum_{i=1}^{N}\alpha_ih_i$$
Here, $\alpha_i$ is the weight/attention of a hidden state $h_i$ calculated by computing Softmax over scores of each hidden state. The score of each individual $h_i$ is calculated by forwarding $h_i$ through a multi-layer perceptron that outputs a score. 

The context vector $\mathbf{c}$ is finally concatenated to the output of the CNN module. Together, this large feature vector is then fed to an MLP which outputs the binary probability distribution of the sentence being sarcastic/non-sarcastic.

\section{Experiments}  \label{sec:sec4}
\subsection{Baseline}
With new dataset in hand, we tweak the model of \cite{amir2016modelling} and consider it as a baseline. We remove the author embedding component because now the sarcasm is independent of authors (it is based on current events and common knowledge). The CNN module remains intact.

\subsection{Experimental Setup}
To represent the words, we use pre-trained embeddings from word2vec model and initialize the missing words uniformly at random in both the models. These are then tuned during the training process. We create train, validation and test set by splitting data randomly in 80:10:10 ratio. We tune the hyper-parameters like learning rate, regularization constant, output channels, filter width, hidden units and dropout fraction using grid search. The model is trained by minimizing the cross entropy error between the predictions and true labels, the gradients with respect to the network parameters are computed with backpropagation and the model weights are updated with the AdaDelta rule. Code for both the methods is available on GitHub\footnote{\url{https://github.com/rishabhmisra/Sarcasm-Detection-using-CNN}}.

\section{Results and Analysis}  \label{sec:sec5}
\subsection{Quantitative Results}
\begin{figure}[h!]
    \includegraphics[width=\linewidth]{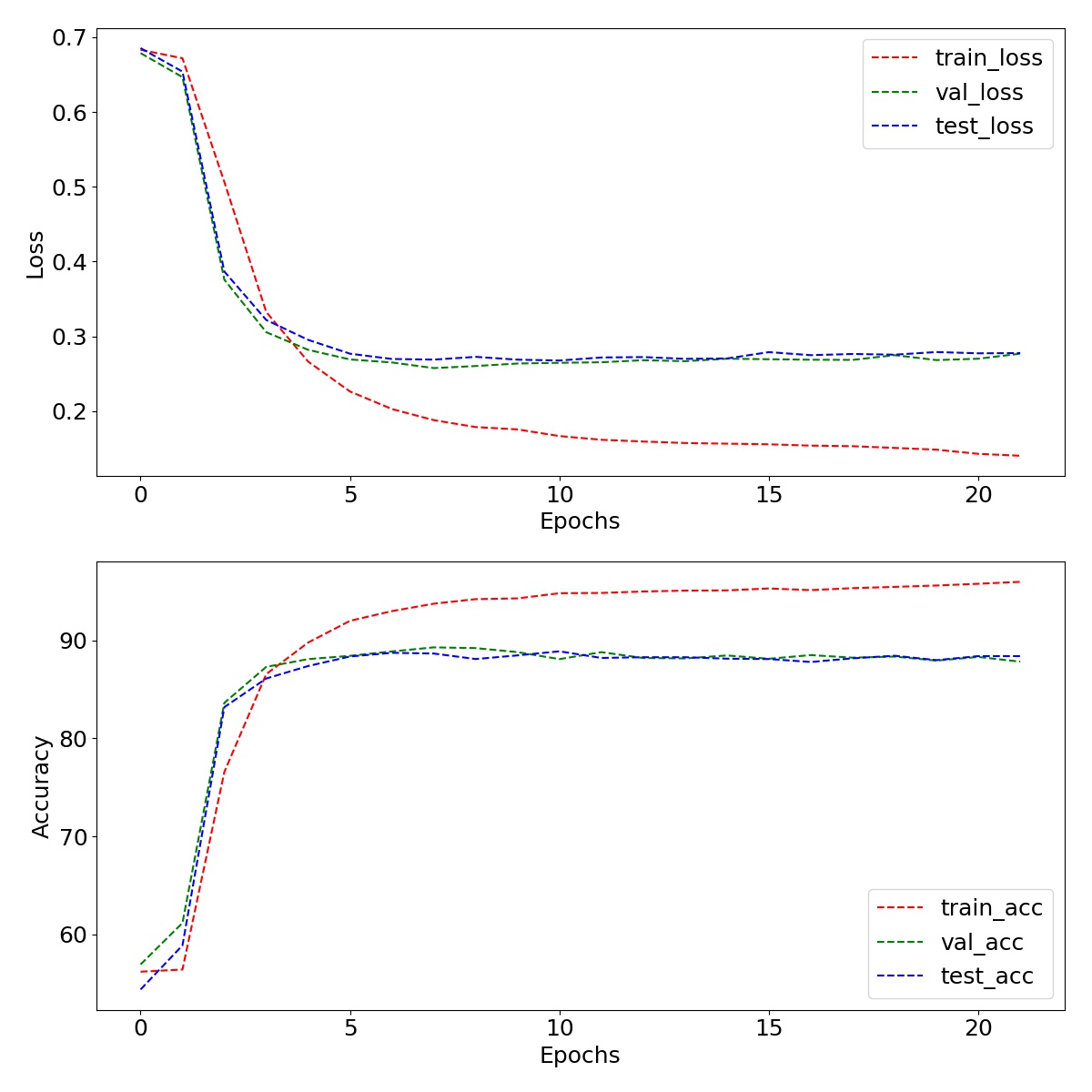}
    \caption{Loss and accuracy trend of the proposed method.}
    \label{fig:fig2}
\end{figure}

We report the quantitative results of the baseline and the proposed method in terms of classification accuracy, since the dataset is mostly balanced. The final classification accuracy after hyper-parameter tuning is provided in \Cref{tab:table2}. As shown, our model improves upon the baseline by $\sim$ 5\% which supports our first hypothesis mentioned in \cref{sec:sec3}. The performance trend of our model is shown in \Cref{fig:fig2}. % TODO: could include performance trend

\begin{table}[h!]
  \centering
  %\resizebox{\columnwidth}{!}{%
  \begin{tabular}{lrr}
  \toprule
  \textbf{Implementation} & \textbf{Test Accuracy} \\
  \midrule
  Baseline & 84.88\% \\
  Proposed method & 89.7\% \\
  \bottomrule
\end{tabular}%}
  \caption{Performance of baseline and proposed method in terms of classification accuracy}
  \label{tab:table2}
\end{table}

\subsection{Qualitative Results}
\begin{figure} %{\linewidth}
    \includegraphics[width=\linewidth]{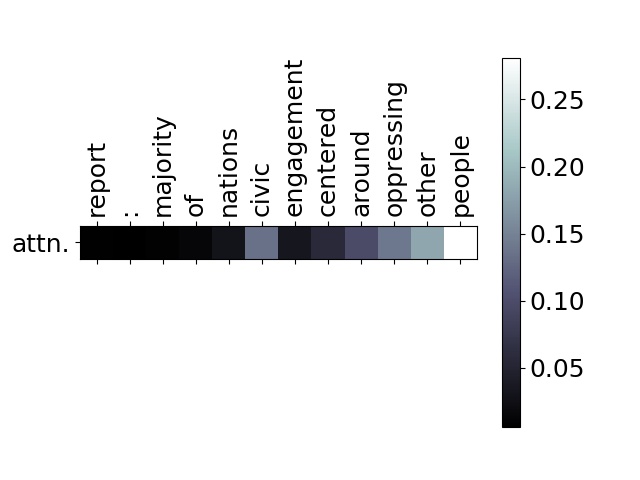}
    \vspace*{-5mm}
    \caption{Attention layer output to showcase co-occurrence of incongruent word phrases}
    \label{fig:fig4}
\end{figure}

\begin{figure} %{\linewidth}
    \includegraphics[width=\linewidth]{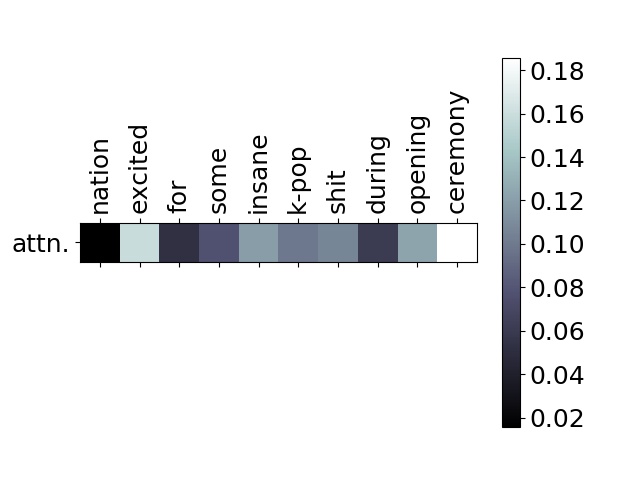}
    \vspace*{-5mm}
    \caption{Attention layer output to showcase co-occurrence of incongruent word phrases}
    \label{fig:fig5}
\end{figure}

\begin{figure} %{\linewidth}
    \includegraphics[width=\linewidth]{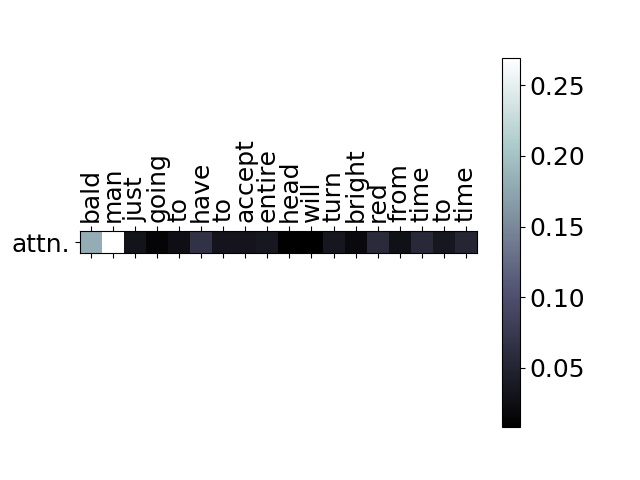}
    \vspace*{-5mm}
    \caption{Attention layer output to showcase insincerity}
    \label{fig:fig6}
\end{figure}

\begin{figure} %{\linewidth}
    \includegraphics[width=\linewidth]{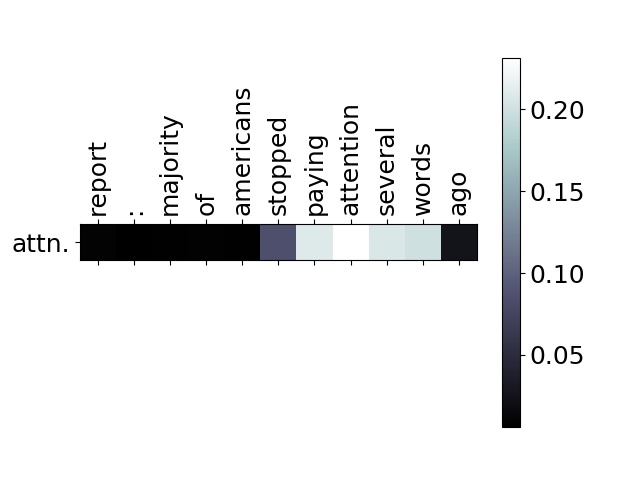}
    \vspace*{-5mm}
    \caption{Attention layer output to showcase satirical nature}
    \label{fig:fig7}
\end{figure}

We visualize the attention over some of the sarcastic sentences in the test set that are correctly classified with high confidence scores. This helps us better understand if our hypothesis is correct and provides better insights into Sarcasm Detection process. \Cref{fig:fig4} and \Cref{fig:fig5} show that the attention module emphasizes on co-occurrence of incongruent word phrases within each sentence, such as `civic engagement' \& `oppressing other people'  in \ref{fig:fig4} and `excited for' \& `insane k-pop sh*t during opening ceremony' in \ref{fig:fig5}. This incongruency is an important cue for us humans too and supports our second hypothesis mentioned in \cref{sec:sec3}. This has been extensively studied in \cite{joshi2015harnessing}. \Cref{fig:fig6} shows that presence of 'bald man' indicates that this news headline is rather insincere probably meant for ridiculing someone. Similarly, `stopped paying attention' in \Cref{fig:fig7} has more probability to show up in satirical sentence, rather than a sincere news headline.

% \begin{figure}
%     \begin{subfigure}[h]{0.49\linewidth}
%     \includegraphics[width=\linewidth]{1_4.jpg}
%     \vspace*{-5mm}
%     \caption{}
%     \label{fig:fig4a}
%     \end{subfigure}
% \hfill
%     \begin{subfigure}[h]{0.49\linewidth}
%     \includegraphics[width=\linewidth]{1_3.jpg}
%     \vspace*{-5mm}
%     \caption{}
%     \label{fig:fig4b}
%     \end{subfigure}
    
%     \begin{subfigure}[h]{0.49\linewidth}
%     \includegraphics[width=\linewidth]{1_2.jpg}
%     \vspace*{-5mm}
%     \caption{}
%     \label{fig:fig4c}
%     \end{subfigure}
% \hfill
%     \begin{subfigure}[h]{0.49\linewidth}
%     \includegraphics[width=\linewidth]{1_1.jpg}
%     \vspace*{-5mm}
%     \caption{}
%     \label{fig:fig4d}
%     \end{subfigure}   
% \caption{Visualizing attention over the entire length of the sarcastic sentences}
% \label{fig:fig4}
% \end{figure}

\section{Future Work}  \label{sec:sec6}
We are left with several unexplored directions that we would like to work on in future. Some of the important directions are as follows:
\begin{itemize}
\item We plan to perform ablation study on our proposed architecture to analyze the contribution of each module.
\item The approach proposed in this work could be considered as a pre-computation step and the learned parameters could be tuned further on Semeval dataset. Our intuition behind this direction is that this pre-computation step would allow us to capture the general cues for sarcasm which would be hard to learn on Semeval dataset alone (given its small size). This type of transfer learning is shown to be effective when limited data is available \cite{pan2010survey}.
\item Lastly, we observe that detection of sarcasm depends a lot on common knowledge (current events and common sense). Thus, we plan to integrate this knowledge in our network so that our model is able to detect sarcasm based on which sentences deviate from common knowledge. Recently, \cite{young2017augmenting} integrated such knowledge in dialogue systems and the ideas mentioned could be adapted in our setting as well.
\end{itemize}

\bibliographystyle{ACM-Reference-Format}
\bibliography{sarcasm_main}

\end{document}